\documentclass{ecai2014}
\usepackage{times}
\usepackage{graphicx}
\usepackage[justification=centering]{caption}
\usepackage{subfigure}
\usepackage{latexsym}
\usepackage{amsfonts}
\usepackage{float}

\ecaisubmission   % inserts page numbers. Use only for submission of paper.
\begin{document}

\title{Off-Policy Shaping Ensembles in Reinforcement Learning}

\author{Anna Harutyunyan \and Tim Brys \and Peter
Vrancx \and Ann Now\'{e}\institute{AI Lab, Vrije Universiteit Brussel,
Belgium, email: \{anna.harutyunyan, timbrys, pvrancx, anowe\}@vub.ac.be} }

% \author{Anna Harutyunyan\footnote{Vrije Universiteit Brussel,
% Belgium, email: \{anna.harutyunyan, tbrys, prvancx, anowe\}@vub.ac.be
% } \and Tim Brys\footnotemark[\value{footnote}] \and Peter
% Vrancx\footnotemark[\value{footnote}]  \and Ann Now\'{e}\footnotemark[\value{footnote}] }

\maketitle
\bibliographystyle{ecai2014}

\begin{abstract}
 Recent advances of gradient temporal-difference
 methods allow to learn off-policy multiple value functions in
 parallel without sacrificing convergence guarantees or computational efficiency. This opens
 up new possibilities for sound ensemble techniques in reinforcement learning.
  In this work we propose learning an ensemble of policies related through
  potential-based shaping rewards. The ensemble induces a combination
  policy by using a voting mechanism on its components. Learning
  happens in real time, and we empirically show the combination policy to
  outperform the individual policies of the ensemble.
\end{abstract}

\section{Introduction}
\label{sec:introduction}

% WHAT WE DID AND WHY IT IS WORTHWHILE.

{\em Reinforcement learning (RL)} is a framework~\cite{sutton-barto98}, where an {\em agent} learns
from interacting with its (typically Markovian) environment. The bulk
of RL algorithms focus on the {\em on-policy} setup, in which the agent learns only about the policy it
is executing. While the {\em
  off-policy} setup, in which the agent's {\em behavior} and {\em
  target} policies are allowed to differ is arguably more versatile,
its use in practice has been hindered by the convergence issues
arising when combined with function approximation (a likely
scenario, given any reasonable problem); e.g. the popular Q-learning
potentially diverges~\cite{baird95}. This issue was recently
resolved by the advancement of the family of {\em gradient
  temporal-difference} methods, such as Greedy-GQ~\cite{maei2010}. An interesting
implication of this is the possibility to learn multiple tasks in parallel from a shared experience
stream in a sound framework, an architecture dubbed Horde by Sutton et
al~\cite{sutton11}. In the spirit of ensemble methods~\cite{wiering08}, we use this idea in the
context of learning a {\em single} task faster. Our larger aim is to
devise ensembles of policies that improve the (off-policy) {\em learning speed} of a task online in real time, without incurring extra sample or computational costs.

% In RL as a framework that is based on environment interactions, learning speed
% is one of the primary desirable metrics. 
% Our primary aim is to
% improve learning speed\footnote{This can be expressed as a combination
%   of RL metrics, {\em jumpstart} and {\em time to threshold}, as proposed Taylor et al.~\cite{???}} without increasing sample complexity. % and better

We choose the policies in our ensemble to be related through {\em potential-based
reward shaping}% , and use rank voting to induce a combination policy
. Reward shaping is a well-known technique to speed up the learning
process by injecting domain knowledge into the reward function.
The idea of considering {\em multiple} shaping signals
instead of a single one, is relatively recent: Devlin et al.
observe that it improves performance in the multi-agent
context~\cite{devlin11}, and Brys et al. using a {\em
  multi-objectivization} formalism demonstrate its usefullness while treating different shapings as
correlated objectives~\cite{brys2013}. %Different shapings enrich the  %TODO: SENTENCE ON BENEFIT OF THIS

% Our aim is to develop a learning ensemble that learns a policy for a
% single task better than an individual, and does so in real time.

The scenario we consider in this paper is that of off-policy learning under fixed
behavior, a scenario Maei et al.~\cite{maei2010} refer to as
{\em latent learning}. This is often the setup in applications where the
environment samples are costly and a failure is highly penalized,
making the usual trial and error tactic implausible, e.g. robotic
applications. One can imagine an agent executing a safe
exploratory policy, while learning control policies for a variety of
tasks. 

We note that even though the effects of reward shaping in this
latent learning context are bound to be limited, since a large part of its
benefits lie in guiding exploration {\em during} learning,
% while we assume no control over the agent's behavior, 
we witness a significant rise in performance, making this a
validation of the effectiveness of reward shaping purely as a means of faster knowledge
propagation. See Section~\ref{sec:shaping-policy} for a discussion.

Unlike the existing ensembles in RL, this is the first policy ensemble
architecture capable of learning online in real-time and sound
w.r.t. convergence in realistic setups -- guarantees provided by
Horde~\cite{sutton11}. The limitation (as with Horde in general) is
that it can only be applied in the latent learning setup, to ensure convergence. 

\paragraph{Outline} In the following section, we give a brief overview of definitions
and notation. Section~\ref{sec:parallel-shaping} further motivates the
use of Horde and multiple shaping signals to form our
ensemble. Section~\ref{sec:our-architecture} summarizes our architecture, and describes the rank voting mechanism used for
combining policies. Section~\ref{sec:experimental-results} gives experimental
results in the mountain car domain, and Section~\ref{sec:concl-future-work}
concludes and discusses future work directions.

\section{Background}
\label{sec:background}

% In this section we give the background necessary for understanding
% this paper.

% An RL agent interacts with the environment at
% discrete time steps $t=0,1,2,\ldots$. 
The environment of a RL agent is usually modeled as a Markov
Decision Process (MDP)~\cite{puterman94} given by a 4-tuple $\langle
S, A, T, R\rangle$, where $S$ is the set of states, $A$ is the set of
actions available to the agent% (in general, this is a function of
% state, $A(s)$)
, $T : S\times A\times S\rightarrow {\mathbb R}$ is the {\em transition} function with
$T(s,a,s')$ denoting the probability of ending up in state $s'$ upon
taking action $a$ in state $s$, and $R: S\times A\times S\rightarrow
{\mathbb R}$ is the {\em reward} function with $R(s,a,s')$ denoting
the expected reward on the transition from $s$ to $s'$ upon
taking action $a$. The Markovian assumption is that $s_{t+1}$ and the 
reward $r_{t+1}$ only depend on $s_t$ and $a_t$, where $t$ denotes the
discrete time step. A stochastic {\em policy} $\pi : S\times A\rightarrow
{\mathbb R}$ defines a probability distribution for actions in each state: 

\begin{equation}
  \label{eq:14}
  \pi(a,s) = Pr(a_t=a|s_t=s)
\end{equation}

{\em Value functions} estimate the utility of policies via their
expected cumulative reward. In the discounted setting, the {\em
  state-action} value function $Q^{\pi} : S\times A\rightarrow {\mathbb R}$
is given by:

\begin{equation}
  \label{eq:9}
  Q^{\pi}(s,a) = E_{\pi} [\gamma^t r_{t+1}+\gamma^{t+1} r_{t+2}+\ldots
  | s_t=s,a_t=a] % E_{\pi}\left[\sum_{s_t=s,a_t=a}^{\infty}\gamma^t r_{t+1}\right],
\end{equation}

where $\gamma\in (0,1]$ is the {\em discounting factor}, and $Q$ is
stored as a table with an entry for each state-action pair.

A policy is {\em optimal} if its value is maximized for all state-action
pairs. Solving an MDP implies finding the optimal policy. When the
environment dynamics (given by $T$ and $R$) are unknown, one can solve
the MDP by applying the family of {\em temporal
  difference (TD)} algorithms~\cite{sutton-barto98} to iteratively
estimate the value functions. The following is the update rule of the
popular Q-learning method in its simplest form~\cite{watkins1992}:

\begin{equation}
  \label{eq:10}
  Q^{\pi}(s_t,a_t) \leftarrow Q^{\pi}(s_t,a_t)+\alpha\delta_t
\end{equation}

\begin{equation}
  \label{eq:13}
  \delta_t=(r_{t+1}+\gamma \max\limits_{a*\in A} Q^{\pi}(s_{t+1},a^*) -Q^{\pi}(s_t,a_t))
\end{equation}

where $r_{t+1}$ is the reward received at the transition
$(s_t,a_t,s_{t+1}$), $\alpha$ is the {\em learning
rate} or step size, $\delta_t$ is the {\em TD error} and $s_{t+1}$ is
drawn according to $T$ given $a_t$. {\em Eligibility traces}
controlled by a {\em trace decay} parameter $\lambda$ can be
used as a way to speed up knowledge
propagation~\cite{sutton-barto98}.

Jaakkola et al.~\cite{Jaakkola94convergenceof} show that in the
tabular case this process converges to the optimal solution, under
standard stochastic approximation assumptions.

When the state or action spaces are too large, or continuous, tabular
representations do not suffice and one needs to use function
approximation (FA). The state (or state-action) space is then represented through a set of features $\phi$, and the algorithms
learn the value of a parameter vector $\theta$. In the (common) linear
case:

\begin{equation}
  \label{eq:5}
  Q(s_t,a_t)=\theta^T\phi_{s_t,a_t}
\end{equation}

and (\ref{eq:10}) becomes:

\begin{equation}
  \label{eq:11}
  \theta_{t+1} \leftarrow \theta_t + \alpha\delta_t\phi_t,
\end{equation}

where we slightly abuse notation by letting $\phi_t$ denote the state-action
features $\phi_{s_t,a_t}$, and $\delta_t$ is still computed according to
(\ref{eq:13}). % , and $\theta_t$ be the parameter vectors $\theta_{a_t}$ for all
% actions $a_t$. 

%TODO: OPTIMAL VALUE FUNCTIONS

%TODO: TD METHODS ALWAYS FIND OPTIMAL POLICIES

%TODO: ELIGIBILITY TRACES

\subsection{Horde} 

Unfortunately, FA can cause off-policy bootstrapping methods, such as
Q-learning, to diverge even on simple problems~\cite{baird95,Tsitsiklis97ananalysis}. % TODO: SENTENCE ON
% WHY.
The family of {\em gradient temporal-difference} (GTD) algorithms
resolve this issue for the first time, while keeping
the constant per-step complexity, provided a fixed (or slowly changing)
behavior~\cite{sutton09,maei2010gq}. They accomplish this\footnote{Please
refer to Maei's dissertation for the full details~\cite{maei-diss}.}
by performing gradient descent on a reformulated objective function,
which ensures convergence to the TD fixpoint by introducing a gradient
bias into the TD update. Mechanistically, it requires maintaining and learning a second set of
weights $w$, along with $\theta$, with the following update
rules:\footnote{This is the simplest form of the update rules for
  gradient temporal-difference algorithms, namely that of TDC~\cite{sutton09}. GQ($\lambda$) augments this update with eligibility traces.}

\begin{eqnarray}
  \label{eq:12}
  \theta_{t+1} & \leftarrow &  \theta_t+\alpha_t\delta_t\phi_t-\alpha\gamma\phi'_t(\phi_t^T
  w_t)\\
  w_{t+1}    & \leftarrow &  w_t+\beta_t(\delta_t-\phi_t^T w_t)\phi_t
\end{eqnarray}

%UPDATE RULES??

Off-policy learning allows one to learn about any policy, regardless
of the behavior policy being followed. One then does not need to limit
themselves to a single policy, and may learn about an arbitrary
number of policies from a single stream of environment interactions
(or {\em experience}), with computational
considerations being the bottleneck. GTD methods not only reliably converge in
realistic setups (with FA), but unlike second order algorithms with
similar guarantees (e.g. LSTD~\cite{Bradtke96linearleast-squares}),
run in constant time and memory per-step, and are hence scalable. % This scalability was the
% missing piece in the puzzle of parallel real-time learning in
% RL.
Sutton et al.~\cite{sutton11} formalize a framework of parallel
real-time off-policy learning, naming it {\em Horde}. They demonstrate
Horde being able to learn a set of predictive and goal-oriented
value functions\footnote{Sutton et al.~\cite{sutton11} give Horde in
terms of {\em general value
functions}, each with 4 auxilary inputs: $\pi,\gamma,r,z$. In this
paper we always assume $\pi$ to be the greedy policy w.r.t. to $Q$,
$\gamma$ and $z$ shared between all demons, and $r$ to be related to
the base reward via a shaping reward.} in real-time from a single unsupervised stream of
sensorimotor experience. There have been further successful
applications of Horde in realistic robotic
setups~\cite{pilarski2013}. We take a different angle to the existing
literature in an attempt to use the power of Horde for learning about a {\em
  single} task from multiple viewpoints. 

% These uses have all focused on learning
% multiple tasks, or information about the world, 

%TODO: PICTURE

%LOOK AT HORDE PAPER

% TODO: HORDE SUMMARY, EMPHASIZE PARALLEL REAL-TIME

 % TODO: DECIDE ABOUT GVFs. For easy of understanding, we just assume
% each demon learning with a GQ($\lambda$) rule. 
% policies (value functions) learnt in the Horde framework as, with
% $D$ denoting a set of demons.

  \subsection{Reward shaping} 
\label{sec:reward-shaping}
{\em Reward shaping} augments the true
  reward signal with an additional heuristic reward, provided by the
  designer. It was originally thought of as a way of scaling up
  RL methods to handle difficult problems~\cite{Dorigo97robotshaping}, as RL generally suffers
  from infeasibly long learning times. If applied
  carelessly, however, shaping can slow down or even prevent finding the optimal policy~\cite{Randløv98}. Ng et al.~\cite{ng99} show that grounding the
  shaping rewards in {\em state potentials} is
  both necessary and sufficient for ensuring preservation of the
  (optimal) policies of the original MDP. {\em Potential-based reward
    shaping} maintains a potential function $\Phi : S \rightarrow
  {\mathbb R}$, and defines the auxiliary reward function $F$ as: 
\begin{equation}\label{eq:6}
  F(s,a,s')=\gamma\Phi(s')-\Phi(s)
\end{equation}

  where $\gamma$ is the main discounting factor. Intuitively,
  potentials are a way to encode the desirability of a state, and the
  shaping reward on a transition signals positive or negative progress
  towards desirable states. Potential-based shaping has
  been repeatedly validated as a way to speed up learning in problems with uninformative
  rewards~\cite{grzes-kudenko2009}. 

% In fact, Wiewiora~\cite{Wiewiora2003} showed that in the tabular case
% potential-based shaping is equivalent to Q-table initializitaion.

We refer to the rewards augmented with shaping signals as {\em
  shaped rewards}, the value functions w.r.t. them as {\em shaped value
  functions}, and the greedy policies induced by the shaped value
functions as {\em shaped policies}. Shaped policies converge to the
same (optimal) policy as the base policy, but differ during the
learning process.

\section{Ensembles of Shapings}
\label{sec:parallel-shaping}

In this section we further motivate why we find Horde to be a
well-suited framework for ensemble learning by surveying ensemble
methods in reinforcement learning, and argue why policies
obtained by potential-based reward shaping are good candidates for
such an ensemble.

Ensemble techniques such as boosting~\cite{freund96adaboost} and
bagging~\cite{breiman1996bagging} are widely used in
supervised learning as effective methods to reduce bias and variance
of solutions. The use of ensembles in RL has been extremely
sparse thus far. Most previous uses of ensembles of policies involved independent runs
for each policy, with the combination happening
post-factum~\cite{fausser2011}. This is limited in practical usage,
since it requires a large computational and sample overhead,
assumes a repeatable setup, and does not improve learning
speed. Others, in general, lack convergence guarantees,\footnote{See the
  discussion on convergence in Section 6.1.2 of van Hasselt's
  dissertation~\cite{hasselt2011}.} either using mixed on- and off-policy
learners~\cite{wiering08}, or Q-learners under function
approximation~\cite{brys2013}. In general, an
off-policy setup seems inevitable when considering ensembles of policies;
it is surely only interesting if the policies reflect information
different from the behavior, since the strength of ensemble learning
lies in the diversity of information its components
contribute~\cite{Krogh95neuralnetwork}. Q-learning in this setup is not reliable in the presence of FA. While the unofficial mantra is that in practice under a sufficiently
similar (e.g. $\epsilon$-greedy) policy, Q-learning used with FA does
not diverge, even despite the famous counterexamples~\cite{baird95,Tsitsiklis97ananalysis},
ensembles of {\em diverse} Q-learners are bound to have larger disagreement amongst
themselves and with the behavior policy, and have a much larger
potential of becoming unstable.\footnote{See the discussion in Section
8.5 of Sutton and Barto~\cite{sutton-barto98} relating potential to
diverge to the proximity of behavior and target policies. To the best
of our knowledge, there have been no formal results on this topic.}

The ability to learn multiple policies reliably in parallel in a
realistic setup is provided by the Horde architecture. For this reason, we
believe Horde to be an ideally suited framework for ensemble learning
in RL.% , and one of the necessary building blocks on the path to enjoying
% the full benefits of ensemble learning in RL.

% Hence, the ability to learn ensembles efficiently and soundly
% in a realistic setup is provided by Horde architecture. We believe
% Horde to be an ideally
% suited framework for ensemble learning, and one of the necessary
% building blocks in enjoying the full benefits of ensemble learning in
% RL
%  and make our first attempt in designing a sound reinforcement
% learning ensemble.

Now we turn to the question of the choice of components of our
ensemble. Recall that our larger aim is to use ensembles to speed up learning of a
single task in real time. Krogh and
Vedelsby~\cite{Krogh95neuralnetwork} show in the context of neural networks that effective ensembles have accurate and diverse
components, namely that they make their errors at different parts of
the space. In the RL context this diversity can be expressed through
several aspects, related to dimensions of the learning process: (1)
diversity of {\em experience}, (2) diversity of {\em algorithms} and (3)
diversity of {\em reward signals}. 
Diversity of experience naturally implies high sample complexity, and
assumes either a multi-agent setup, or learning in stages. Diversity of algorithms
may run into convergence issues, unless all algorithms are sound
off-policy, by the argument above. Marivate and
Littman~\cite{marivate2013} consider diversity of {\em MDPs}, by
improving performance in a generalized MDP through an ensemble trained
on sample MDPs, which also requires a two-stage learning process. In the
context of our aim of improving learning speed, we focus on the latter
aspect of diversity: diversity of reward signals.

% \begin{enumerate}
% \item Diversity of experience
% \item Diversity of algorithms
% \item Diversity of reward signals
% \end{enumerate}

%TODO: REWRITE NEXT TWO PARAGRAPHS

As discussed in Section~\ref{sec:reward-shaping}, potential-based
reward shaping provides a framework for enriching the base reward by
incorporating heuristics that express the desirability of states. One can usually think of multiple
such heuristics for a single problem, each effective in different
situations. Combining them na\"\i vely, e.g. with linear
scalarization on the potentials, may be uninformative since the heuristics may counterweigh each
other at some parts of the space, and ``cancel out''. On the other
hand, it is typically infeasible for the designer to handcode all
tradeoffs without executing each shaping separately. Horde provides a
sound framework to learn and maintain {\em all} of the shapings in
parallel, enabling the possibility of using any (scale free) ensemble methods for combination.

% A solution proposed by Brys et al.~\cite{brys2013} is to
% maintain {\em all} of the shapings at hand. Horde provides the sound
% framework to do so in real time, and we suggest using (scale-free) ensemble methods
% for combination

% Most previous research has focused
% on either defining a potential function based on a single heuristic.
% or combining several heuristics into a single potential via an a
% priori linear combination.~\cite{???}.
% As with most dimensionality reduction
% techniques, this incurs an information loss~\cite{???}, as different
% potentials may counterweigh each other at some part of the state
% space. 

\paragraph{Shaping off-policy}
\label{sec:shaping-policy}

We note that we are straying from convention in using reward
shaping in an off-policy latent learning setup. The effects of reward
shaping on the learning process are usually considered to lie in the
guidance of exploration during
learning~\cite{grzes2010diss,marthi2007,ng99}. Laud and
DeJong~\cite{laud03} formalize this by showing that the difficulty of
learning is most dependent on the {\em reward horizon}, a measure of the
number of decisions a learning agent must make before experiencing
accurate feedback, and that reward shaping artificially reduces this
horizon. In our setting we assume no control over the agent's
behavior, and the performance benefits in
Section~\ref{sec:experimental-results} must be explained by a
different effect. Namely, shaping rewards in the TD updates aid faster
{\em knowledge propagation}, which we now observe decoupled from
guidance of exploration due to the off-policy latent learning setup.

In the next section we describe the exact architecture used for this
paper, and the combination method we chose.

\section{Architecture}
\label{sec:our-architecture}

% We are now ready to describe the architecture we propose for real-time
% ensemble learning.

We maintain our Horde of shapings as a set $D$ of
Greedy-GQ($\lambda$)-learners. The reward function is a vector: ${\bf
  R}=\langle R+F_0,R+F_1,\ldots,R+F_{|D|-1}\rangle$, where $F_0=0$
  ($d_0$ always learns on the base reward alone), and $F_i$,
  $i=1,\ldots,|D|-1$ are
potential-based rewards given by (\ref{eq:6}) on potentials $\Phi_1,\Phi_2,\ldots$
provided by the designer. We adopt the terminology of Sutton et
al.~\cite{sutton11}, and refer to individual agents within Horde as {\em demons}.
 Each demon learns a greedy policy $\pi_i$ w.r.t. its
reward $R_i$. We refer to the demons learning on shaped rewards as
shaped demons.

At any point of learning, we can devise a combination policy by collecting
votes on action preferences from all shaped demons
($d_1,d_2,\ldots$). Wiering et al.~\cite{wiering08} discuss several
intuitive ways to do so, e.g. majority voting, rank voting, Boltzman
multiplication, etc. We describe {\em rank voting} used in this paper, but
in general the choice of ensemble combination is up to the designer,
and may depend on the specifics of the problem and architecture. Even
though the base demon $d_0$ does not contribute a vote, we maintain it as a part of the ensemble.

\begin{figure}[h]
  \centering
  \includegraphics[scale=0.5]{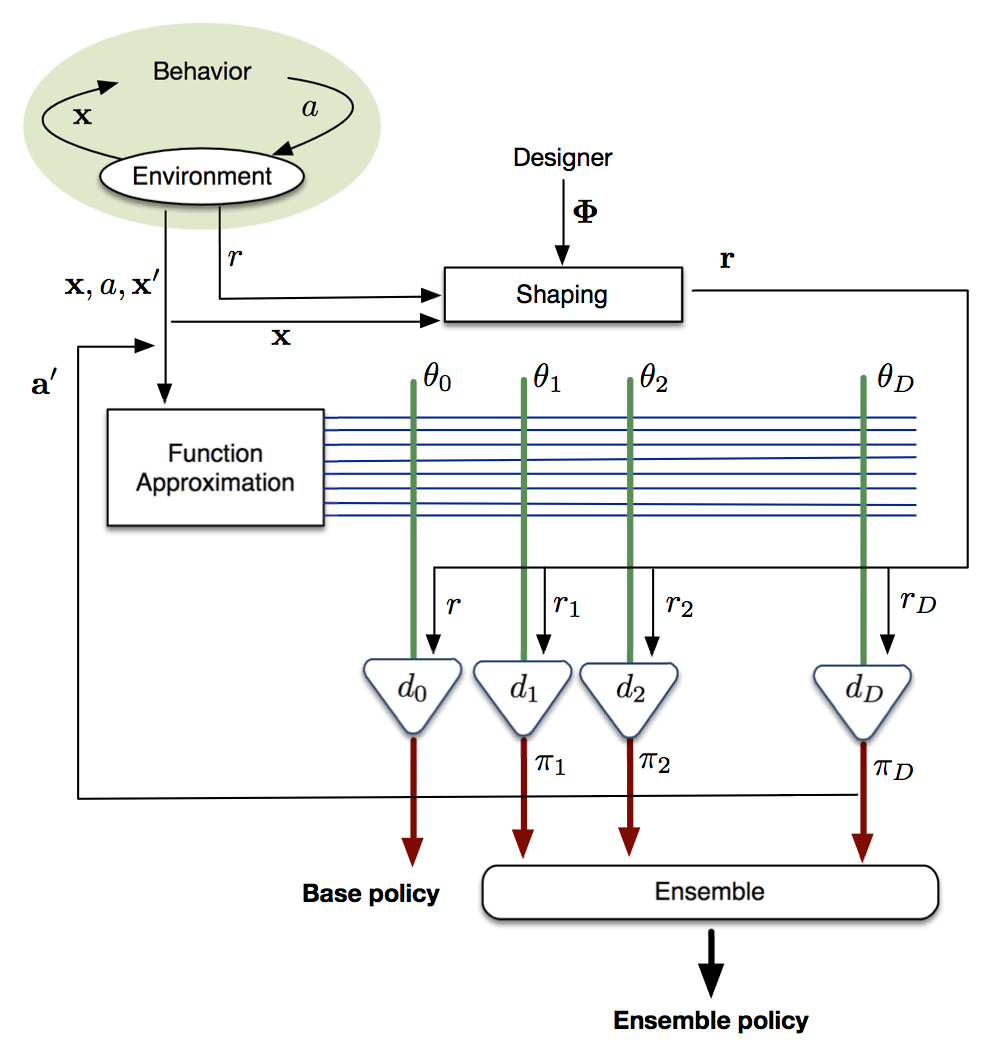}
  \caption{A rough overview of the Horde architecture used to learn an ensemble
    of shapings. The blue output of
    the linear function approximation block are the features of the
    transition (two state-action pairs), with their intersections with 
    $\theta_i$ representing weights. ${\bf a'}$ is
    a vector of greedy actions at ${\bf x'}$ w.r.t. to each policy
    $\pi_i$. Note that all interactions with the environment happen only in the upper left corner.}
  \label{fig:architecture}
\end{figure}

\paragraph{Rank voting} 
Each demon (except for $d_0$) ranks its $n$ actions according to its greedy policy, casting a
vote of $n-1$ for its most, and a vote of $0$ for its least preferred
actions. The voting schema then is defined for policies, rather than
value functions, which mitigates the
magnitude bias.\footnote{Note that even though the
  shaped {\em policies} are the same upon convergence -- the value
  functions are not. } We slightly modify the formulation
from~\cite{wiering08}, by ranking Q-values, instead of policy
probabilities, i.e. let $r: D\times A\rightarrow {\mathbb N}$ be the ranking
map of a demon. Then $r_d(a) > r_d(a')$, if and only if $Q_d(s,a) >
Q_d(s,a')$. The {\em combination} or {\em ensemble} policy acts greedily w.r.t. the cumulative {\em preference} values $P$:

\begin{equation}
  \label{eq:1}
  P(s_t,a)=\sum_{d=1}^{|D|-1} r_d(a),\forall a\in A
\end{equation}

%PREFERENCE VALUES

In the next section we validate our approach on the typical mountain
car benchmark and interpret the results.

% RECENTLY MANY SHAPINGS BLAH BLAH. WHEN WITH FA, FIRST TIME WHEN IT IS
% POSSIBLY TO LEARN ALL SOUNDLY

%COMBINING INTELLIGENTLY IMPORTANT

% IN THE GQ SETTING, WHERE EVERYONE LEARNS IN PARALLEL, UNTAPPED
% POTENTIAL OF LEARNING COMBINATION TECHNIQUES, FITNESS, ETC

% IN THIS PAPER WE USE A FORM OF VOTING IN THE SPIRIT OF WIERING TO
% DEVISE A POLICY AD HOC FROM THE VALUE FUNCTIONS. 

%DESCRIBE VOTING SYSTEM, SCALING IS IMPORTANT

\section{Experiments}
\label{sec:experimental-results}

In this section we give comparison results between the individuals in
our ensemble, and the combination policy. We remind the reader that
while all policies eventually arrive at the same (optimal) solution,
our focus is the time it takes them to get there.

\begin{figure}[h]
  \centering
  \includegraphics[scale=0.4]{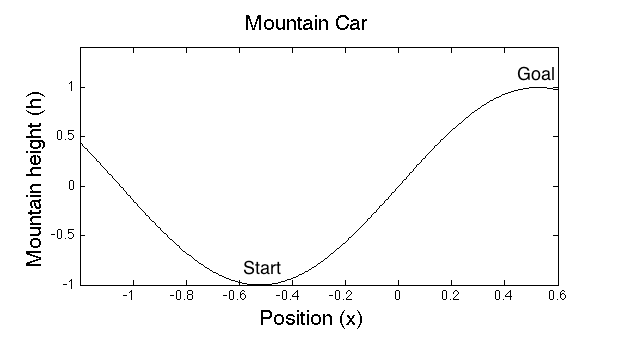}
  \caption{The mountain car problem. The mountain height $h$ is given
    by  $h=\sin (3x)$. % Figure from Taylor's dissertation~\cite{thesis-taylor}. 
    % TODO: ADD CAR, FLAG
  }
  \label{fig:mc}
\end{figure}

We focus our attention to a classical benchmark domain of mountain
car~\cite{sutton-barto98}. The task is to drive an underpowered car up a hill (Fig.~\ref{fig:mc}). %which requires to gain
%momentum on the hill opposite to the target first. 
The (continuous) state of the system is composed of the current position (in $[-1.2,0.6]$)
and the current velocity (in $[-0.07,0.07]$) of the car. Actions are discrete,
a throttle of $\{-1,0,1\}$. The agent starts at the position $-0.5$
and a velocity of $0$, and the
goal is at the position $0.6$. The rewards are $-1$ for every time
step. An episode ends when the goal is reached, or when 2000
steps\footnote{Note the significantly shorter lifetime of an episode here, as compared
  to results in Degris et al.~\cite{degris2012}; since the shaped rewards are more
  informative, they can get by with very rarely reaching the goal.}
have elapsed. The state space is approximated with the standard
tile-coding technique~\cite{sutton-barto98}, using ten tilings of
$10\times 10$, with a parameter vector learnt for each action. The
behavior policy is a uniform distribution over all actions at each time step.

In this domain we define three intuitive shaping potentials. Each is
normalized into the range $[0,1]$.

\begin{description}
\item[Right shaping.] Encourage progress to the right (in the direction of
  the goal). This potential is flawed by design, since in order to
  get to the goal, one needs to first move away from it.
  \begin{equation}
\label{eq:4}
\Phi_1({\bf x})=c_r\times x
\end{equation}
\item[Height shaping.] Encourage higher positions (potential
  energy), where height $h$ is computed according to the formula in Fig.~\ref{fig:mc}.
  \begin{equation}
    \label{eq:2}
    \Phi_2({\bf x})=c_h\times h
  \end{equation}
\item[Speed shaping.] Encourage higher speeds (kinetic energy).
  \begin{equation}
    \label{eq:3}
    \Phi_3({\bf x})=c_s\times |\dot{x}|^2
  \end{equation}
\end{description}

Here ${\bf x} = \langle x,\dot{x}\rangle$ is the state (position and velocity), and ${\bf c} =
\langle c_r,c_h,c_s\rangle$ is a vector of tuned scaling
constants.\footnote{The scaling of potentials is in general a
  challenging problem in reward shaping research. Finding the right
  scaling factor requires a lot of a priori tuning, and the factor is
  generally assumed
  constant over the state space. The scalable nature
  of Horde could be used to lift this problem, by learning multiple
  preset scales for each potential, and combining them via either a
  voting method like the one described here, or a meta-learner. See
  Section~\ref{sec:concl-future-work}.}

Thus our architecture has 4 demons: $<d_0,d_1,d_2,d_3>$, where $d_0$
learns on the base reward, and the others on their respective shaping rewards. The
combination policy is formed via rank voting, which we found to
outperform majority voting, and a variant of Q-value voting on this problem.
% , and only collects votes
% from shaped demons. 
% For a comparison with
% other voting schemas, see Table~\ref{???} in the Appendix BLAH?

%  Each demon learns according to the
% Greedy-GQ($\lambda$) rule, and the shaped demons are combined using
% rank voting.
% We tried majority
% voting, and scaled Q-voting as alternative combination techniques and
% found rank voting to perform best (Table). 

The third (speed) shaping turns out to be the most helpful universally. If
this is the case one would likely prefer to just use that
single shaping on its own, but we assume such information is not
available a priori, which is a more realistic (and challenging) situation. To make our
experiment more interesting we consider two scenarios: with and
without this best shaping. Ideally we would
like our combination method to be able to outperform the two
comparable shapings in the first scenario, and pick out the best
shaping in the second scenario.

We used $\gamma=0.99$. The learning parameters were tuned and selected
to be $\lambda=0.4,\beta=0.0001,\alpha=\langle 0.1,0.05,0.1,0.1\rangle$,
where $\lambda$ is the trace decay parameter, $\beta$ the step size for the
second set of weights in Greedy-GQ, and $\alpha$ the vector of step
sizes for the value
functions of our demons.\footnote{These were tuned individually, as the
value functions differ in magnitude. % Using a single $\alpha=0.1$
% across all demons resulted in decent performance. 
% We conjecture given potentials 
} % The
% combination only collects votes from the shaped policies of the ensemble.
We ran 1000 independent runs of 100 episodes each. The evaluation was
done by interrupting the off-policy learner every 5 episodes, and
executing each demon's greedy policy once. % CONSIDER AVERAGE??
No learning was allowed during evaluation. The graphs reflect the
average base reward. The initial and final performance refer to the
first and last 20\% of a run.

\begin{figure}[h!]
   %\centering
   \subfigure[Scenario 1, with two comparable shapings. The
   combination is able to follow the right shaping in the 
   beginning (where it is best), then switch to the height shaping.]{\includegraphics[scale=0.4]{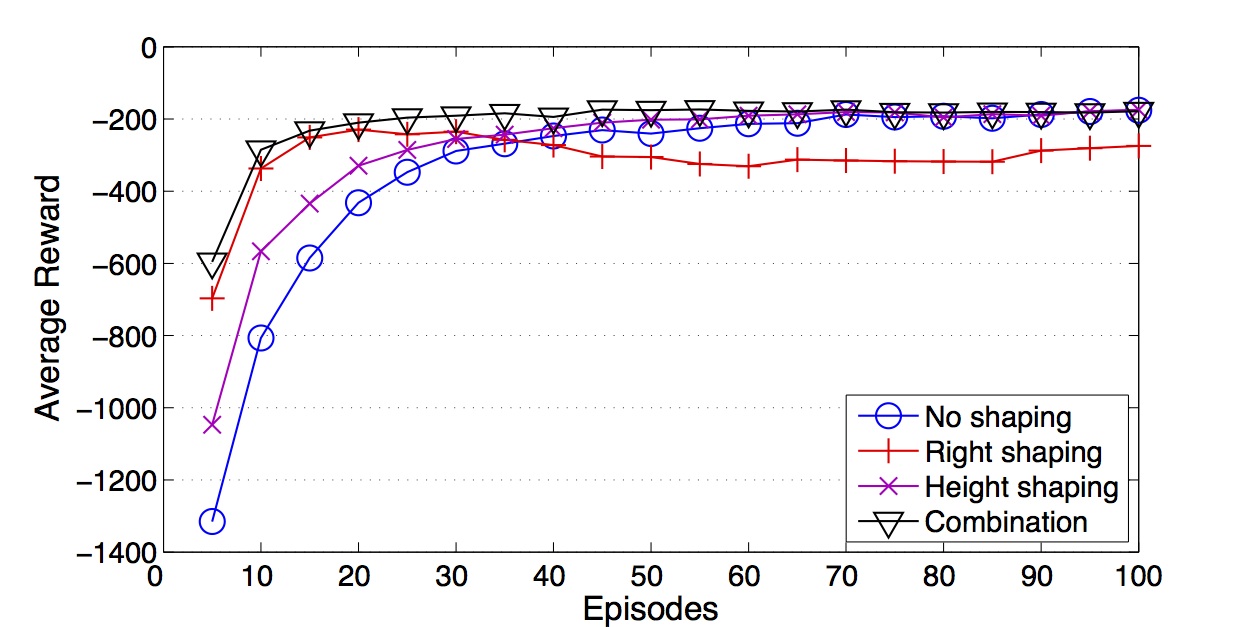}}\label{fig:sc1}
   \subfigure[Scenario 2, with one clearly superior shaping. The
   combination is able to pick out the best shaping and follow it.]{\includegraphics[scale=0.4]{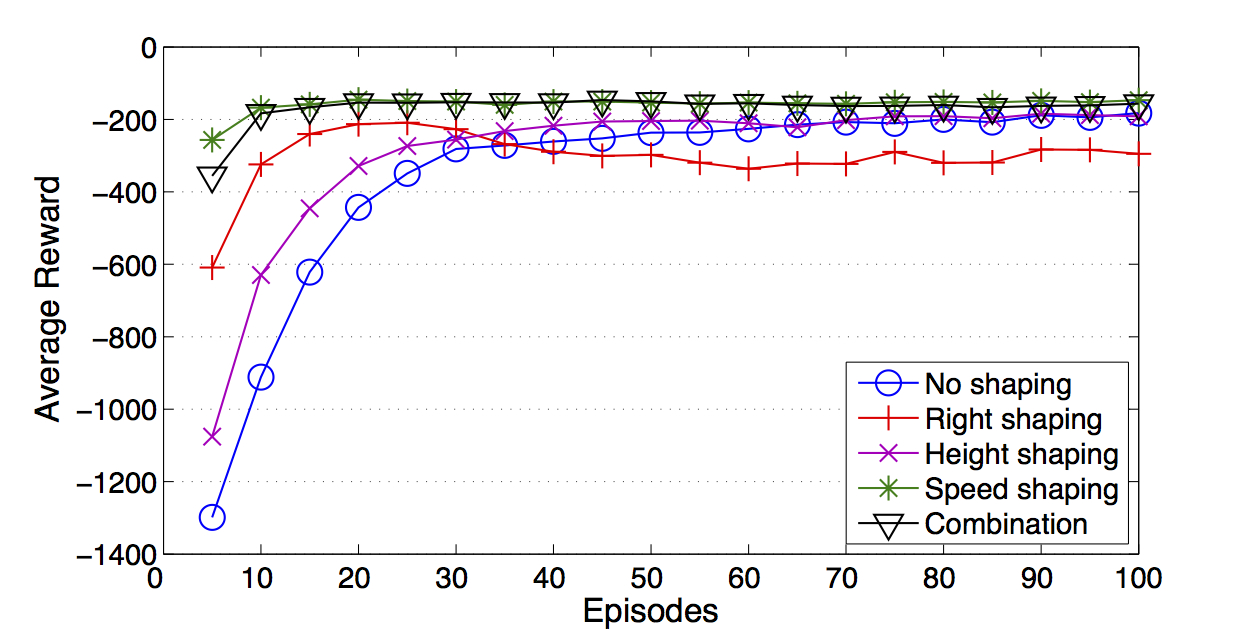}}\label{fig:sc2}
  \caption{Learning curves of the policies in the ensemble in mountain car}\label{fig:learning-curves}
\end{figure}

\begin{table}
\begin{center}
{\caption{Results for the scenario with two comparable shapings. The
    combination has the best cumulative performance. In the initial
    stage it is comparable to the right shaping, in the
    final -- to the height shaping (each being the best in
    the corresponding stages), overall outperforming both. The results
    that are not
    significantly different from the best (Student's t-test with $p>0.05$) are in bold.}\label{tab:sc1}}
\begin{tabular}{|  l | c  c  c |}
\hline
 & \multicolumn{3}{c |}{Performance} \\
  Variant & Cumulative & Initial & Final \\ \hline
  No shaping & -336.3 $\pm$ 279.5  & -784.7 $\pm$ 385.9 &
  -185.1 $\pm$ 9.9 \\
  Right shaping & -310.4 $\pm$ 96.9  & {\bf -378.5 $\pm$ 217.4} &
  -290.3 $\pm$ 19.3 \\
  Height shaping & -283.2 $\pm$ 205.2 & -594.2 $\pm$ 317.0 &
   {\bf -182.3 $\pm$ 7.5} \\
  Combination & {\bf -211.2 $\pm$ 94.2}  & {\bf -330.6 $\pm$ 179.5} &
  {\bf -180.2 $\pm$ 1.5} \\ 
\hline
\end{tabular}
\end{center}
\end{table}

\begin{table}
\begin{center}
{\caption{Results for the scenario with one clearly superior shaping. The
    combination has comparable performance to that shaping, indicating
    that even in such a setup, our technique is viable. The results
    that are not
    significantly different from the best (Student's t-test with $p>0.05$) are in bold.}\label{tab:sc2}}
\begin{tabular}{|  l | c  c  c |}
\hline
 & \multicolumn{3}{c |}{Performance} \\
  Variant & Cumulative & Initial & Final \\ \hline
  No shaping & -349.7 $\pm$ 285.2 & -818.6 $\pm$ 373.7 &
  -193.2 $\pm$ 10.9 \\
  Right shaping & -303.4 $\pm$ 81.4  & -346.7 $\pm$ 181.2 &
  -295.1 $\pm$ 16.7 \\
  Height shaping & -292.4 $\pm$  213.8  & -619.8 $\pm$ 328.3 &
    -190.1 $\pm$ 5.3 \\
  Speed shaping & {\bf -158.6 $\pm$ 23.7} &  {\bf-182.1 $\pm$ 50.6} &
    {\bf-150.2 $\pm$ 2.9} \\
  Combination & {\bf -168.7 $\pm$  44.7}  &  {\bf -214.8 $\pm$ 94.8} &
   -161.7 $\pm$ 4.0 \\ \hline
\end{tabular}
\end{center}
\end{table}

%INTERPRET RESULTS

The results in Fig.~\ref{fig:learning-curves}, and
Tables~\ref{tab:sc1} and \ref{tab:sc2} show that individual shapings
alone aid learning speed significantly. The combination method meets
our desiderata: it either statistically matches
or is better than the best shaping at any stage, overall outperforming
all single shapings. The exception is the final performance of the run in Scenario
2, where the performance of the best shaping is significantly
different from the combination. The difference in actual averaged performance
however is relatively small, and arguably negligible.

% TODO: ALL SHAPINGS CONVERGE TO THE SAME. COMB $>$ SHAPINGS $>$ BASE
% W.R.T. JUMPSTART AND TIME TO THRESHOLD METRICS~\cite{Thesis-taylor}.

We note that even the best performances in these tables do not
reach the maximum attainable, if behaving online.\footnote{An artefact
of value-function methods learnt off-policy under a behavior policy
rarely reaching the goal. % : all demons produce policies with an extra
% ``swing'' in the context of mountain car, with the true optimal region
% remaining relatively unexplored for longer. 
Given longer learning periods, they will
get closer and closer to the attainable optimum, but we choose not to
concern ourselves with this in the context of this paper, as our main
focus lies in improving on the learning time {\em within} the off-policy framework.}

\section{Conclusions and future work}
\label{sec:concl-future-work}

We gave the first policy ensemble that is both sound and
capable of learning in real time, by exploiting the power of Horde
architecture to learn a single policy well. 
The value functions in our ensemble learned on shaped
rewards, and we used a voting method to combine them. We
validated the approach on the classical mountain car domain,
considering two scenarios: with and without a clearly best shaping
signal. In the former scenario, the combination outperformed single
shapings, and in the latter was able to match the performance of that
best shaping. % IT OUTPERFORMED AND KICKED ASS BLAH.
In general, we expect to see larger benefits on larger problems; a
more extensive suite of experiments is subject to future work.

The primary limitation of Horde is the requirement to keep the
behavior policy fixed (or change it slowly). While this is an
important case, relaxing this constraint would further expand the
effectiveness of the architecture. This is a topic of ongoing research in the GTD community.

%TODO: ARGUE THAT IT'S OK, OR THAT USEFUL ANYWAY 

\paragraph{Future work}
\label{sec:future-work}

In this work, we considered an ad-hoc voting approach to combining
shapings. One of the possible future directions would be to {\em learn}
optimal combination ways via predicting some shared fitness value
w.r.t. the policies induced by the learnt value functions. The
challenge with this is that the meta-learning has to happen at a much
faster pace for it to be useful in speeding up the main learning
process. In the case of shapings, this is doubly the case, since they
all eventually converge to the same (optimal) policy. The size of this window
of opportunity is related to the size of the problem.

The scalability of Horde allows for learning potentially thousands of value functions
efficiently in parallel. While in the context of shaping it will
rarely be sensible to actually define thousands of distinct shapings,
one could imagine defining shaping potentials with many different
scaling factors each, and having a demon combining the shapings from
each group. This would not only mitigate the scaling problem, but potentially make the
representation more flexible by having non-static scaling factors
throughout the state space. This has a roughly similar flavor to the approach
of Marivate and Littman~\cite{marivate2013}, who learn to solve many
variants of a problem for the best parameter settings in a generalized MDP.

One could go further and attempt to {\em
  learn} the best potential functions~\cite{marthi2007,grzes2010}. As
before, one needs to be realistic about attainability of learning this in
time, since as argued by Ng et al.~\cite{ng99}, the best potential
function correlates with the optimal value function $V^*$, learning
which would solve the base problem itself and render the potentials
pointless.

% Another avenue to investigate is the choice of the components of the ensemble
% themselves: shaped policies are attractive, since they preserve
% optimality, and the method of combination we considered in this paper
% does not guarantee it in general, but the real strength of true off-policy learning is the
% ability to learn arbitrary pieces of information, or solve arbitrary
% tasks, which then may be combined in an optimal fashion. This
% direction has a flavor of hierarchical methods.

\ack  Anna Harutyunyan is supported by the IWT-SBO
project MIRAD (grant nr. 120057). Tim Brys is funded by a Ph.D grant of the Research Foundation-Flanders (FWO).

\bibliography{all}
\end{document}